# ChatGPT-generated texts show authorship traits that identify them as non-human

V. Dentella[1*], W. Huang[2], S.A. Mansi[1], J. Grieve[2] & E. Leivada[3,4]

[1] Università degli Studi di Pavia, Pavia, Italy
[2] University of Birmingham, Birmingham, UK
[3] Universitat Autònoma de Barcelona, Barcelona, Spain
[4] Institució Catalana de Recerca i Estudis Avançats (ICREA)

[*]correspondence: vittoria.dentella@unipv.it

## Abstract

Large Language Models can emulate different writing styles, ranging from composing poetry that appears indistinguishable from that of famous poets to using slang that can convince people that they are chatting with a human online. While differences in style may not always be visible to the untrained eye, we can generally distinguish the writing of different people, like a linguistic fingerprint. This work examines whether a language model can also be linked to a specific fingerprint. Through stylometric and multidimensional register analyses, we compare human-authored and model-authored texts from different registers. We find that the model can successfully adapt its style depending on whether it is prompted to produce a Wikipedia entry vs. a college essay, but not in a way that makes it indistinguishable from humans. Concretely, the model shows more limited variation when producing outputs in different registers. Our results suggest that the model prefers nouns to verbs, thus showing a distinct linguistic backbone from humans, who tend to anchor language in the highly grammaticalized dimensions of tense, aspect, and mood. It is possible that the more complex domains of grammar reflect a mode of thought unique to humans, thus acting as a litmus test for Artificial Intelligence.

# Introduction

Scholars from different disciplines have been addressing the question of what makes us human for centuries. For Nobel laureate Bertrand Russell, the answer is language, for "no matter how eloquently a dog may bark, he cannot tell you that his parents were poor but honest". Human language is both flexible and constrained at the same time, and this is why the Turing Test, described as a litmus test for Artificial Intelligence [Shieber 1994, French 2000], is linked to achieving a level of conversational proficiency that is highly complex, akin to that of a human [Turing 1950].

Human language is flexible in the sense that we all make different choices when conversing. Every human is thought to have a distinct linguistic fingerprint called *idiolect* [Halliday et al. 1964, Coulthard 2004]. This idiolect, which can be defined as an individual's unique use of linguistic forms (including lexical choices, collocations and fixed expressions, punctuation patterns, misspellings, and grammatical style), is critical for authorship attribution in a range of situations: from identifying that a poem with dashes, elliptical syntax, and unconventional capitalization is more likely authored by Emily Dickinson and not by William Shakespeare, to pinning down a person of interest in the course of a criminal investigation, as happened in the Unabomber case.

At the same time, human language is constrained by the variation it permits depending on its function in each situation. We all adapt our idiolects depending on communicative context: If you are meeting friends, "hey guys!" would be acceptable as a greeting, but if you find yourself in a court room "good morning, Your Honor" would be a more appropriate alternative. Although both sentences convey the same message, they belong to different *registers*. The notion of register can be defined as a subset of specialized language shaped by specific situational features, linguistic functions, social contexts, and communicative purposes [Biber & Conrad 2001, 2019].

Large Language Models (LLMs) manipulate linguistic form in a way that allows for successful completion of a range of Natural Language Processing tasks [Srivastava et al. 2023]. However, the issue of whether artificial systems understand the *meaning* associated with the form they operate on has given rise to an ongoing debate. On one side, following the symbolic theory of the mind, a system capable of manipulating form in a way that is indistinguishable from humans would be capable of understanding meaning [Turing 1950, Weizenbaum 1965, Winograd 1973]. On the other side stands the symbol grounding problem [Harnad 1987, 1990], whereby meaning is understood as being intrinsically related to the cognitive states of those who assign it and thus cannot be inferred from the manipulation of form alone [Searle 1980, Bender & Koller 2020]. Depending on one's viewpoint, conclusions as to how well the current generation of LLMs fares in the Turing test differ, reflecting a tension over a possible distinction between LLMs acquiring mastery of meaning through form vs. manipulating form alone.

On the one hand, LLMs undeniably have the impressive ability to emulate different writing styles. This has recently led to claims that LLMs can write poetry that looks indistinguishable from that of famous poets such as Emily Dickinson [Porter & Machery 2024]. Similarly, GPT-4 performed so well in conversation that it was judged to be a human 54% of the time [Jones & Bergen 2024]. Fine-tuning LLMs to match the idiolect of a specific user is one of the methods currently explored to ensure better alignment between models and their users [Liu et al. 2024].

On the other hand, we find Lady Lovelace's Objection [Turing 1950]: Artificial Intelligence (AI) can do whatever we know how to order it to perform. Put another way, LLMs do not pass the Turing Test because they have developed the ability to read the context, to form beliefs about their interlocutor's communicative intent (i.e. the purpose or motive behind a linguistic message), and then to successfully adapt their language depending on the register, but because, when provided with very precise instructions and possibly even actual chunks of the language they should include in their conversational contribution (e.g., answer with slang such as 'bruh', use misspellings or alternative spellings such as 'smol' for 'small', capitalize most nouns regardless of whether they are proper names or not, combine the frequent use of dashes with vivid imagery and syntactic ellipsis), they can successfully embed those features in their replies, and thus convincingly imitate the writing style of different groups or individuals.

This ability may produce subtle differences in the writing style that may not always be visible to the untrained eye. Not being able to tell apart that "Tell all the truth but tell it slant — Success in Circuit lies" is a line from Dickinson's poetry but "The wind — it bends the Amber stalks — With fingers soft and Light" is not, does not entail that no difference exists, similar to how one's inability to tell apart an authentic Dalí painting from a fake one does not mean that the differences are not there or that they are imperceptible to all. It has been found that AI-generated text can be told apart from human text through diversified methodologies, including human raters, automatic detectors [Ippolito et al. 2019, Verma et al. 2024], and quantitative linguistics measurements [Huang et al. 2024, Uchendu et al. 2020, Munir et al. 2021].

Although detecting differences between human- and model-authored texts is important [Heikkilä 2022, Spitale et al. 2023], this type of research is not always concerned with or capable of providing explanations for *why* human and AI texts differ, and *in what sense*. Does a model like ChatGPT have a unique idiolect too, and if so, can this idiolect be adapted to different registers, thus capturing a hallmark characteristic of human language? In this respect, stylometry has the potential to not only identify differences between human and AI writings, but to also explain them. In this Article, we thus employ stylometry, defined as the quantitative analysis of stylistic variation, used primarily to resolve cases of disputed authorship [Grieve 2023], to comparatively examine whether ChatGPT-3.5 (henceforth, ChatGPT) can (i) be linked to a stable idiolect that (ii) permits register variation. Concisely, register variation is a proxy for communicative intent and stylometry represents a vehicle for assessing whether AI displays any.

The rationale behind this study originates from two linguistic facts. First, linguistic form varies as a function of meaning. In this work, we define understanding meaning as the process of retrieving communicative intent from a linguistic expression [Bender & Koller 2020]. Humans manipulate linguistic features (i.e. form) to best convey an intended meaning [Hymes 1966, 1974]. For instance, the expression of viewpoints during a conversation is mediated by the frequent use of the pronouns *I* and *we*. Register variability indexes sensitivity to context and intentionality behind the transmission of meaning [van Dijk 2005, 2008]. Second, variation in stylometric features can generally be explained by differences between how individual registers are realized by specific people [Grieve 2023]. If a stylometric analysis finds differences in the rate of use of different stylometric features in two corpora, this will generally reflect register variation between these corpora and, consequently, sensitivity to meaning and communicative demands on the generating agent's part (human or artificial).

The degree to which LLMs possess communicative competence and can adapt to register —especially in the *absence* of targeted training and fine-tuning that gives chunks of the appropriate answers to the model— has important consequences for the substitutability of human texts with automatically generated ones [Agnew et al. 2024]. Most importantly, it has implications for determining what LLMs can actually do: Finding evidence for the inability (or limited ability) of LLMs to adapt for register would amount to proof of a more mechanistic —rather than creative, as per Lady Lovelace's Objection— implementation of language, marking an important difference from humans. Crucially, such a result would also provide important empirical evidence for policy makers tasked with developing regulations related to the use of generative AI in education and other sectors [Adams 2024]. Additionally, the mass-scale deployment of machines that generate standardized, inflexible outputs has the potential of driving diachronic evolutive processes, at least in the written domain. For example, academic writing over centuries has become more succinct to allow for a more efficient reporting of information [Biber & Gray 2016] —a shift that the widespread circulation of inflexible, non-adaptive, context-insensitive "AI-idiolects" could undermine.

To assess the degree to which LLMs (i) can be linked to a unique idiolect that (ii) is sensitive to meaning such that it can successfully adapt to the different semantic-pragmatic requirements of different registers, we evaluate one corpus of texts written by humans and one generated by ChatGPT, first through a stylometric Principal Component Analysis of function word frequencies (FW-PCA) [Binongo 2003, Grieve 2023], and then through a Multidimensional Register Analysis (MDA) [Biber 1988, Grieve 2023]. Our Research Questions (RQs) are the following: **RQ1**. Are there systematic stylistic differences between the idiolects of humans and ChatGPT regardless of register? How much can these differences be controlled via prompting? **RQ2**. Are there systematic stylistic differences between humans and ChatGPT in how variation across registers is realized? Figure 1 summarizes these RQs and the overall aim of this work.

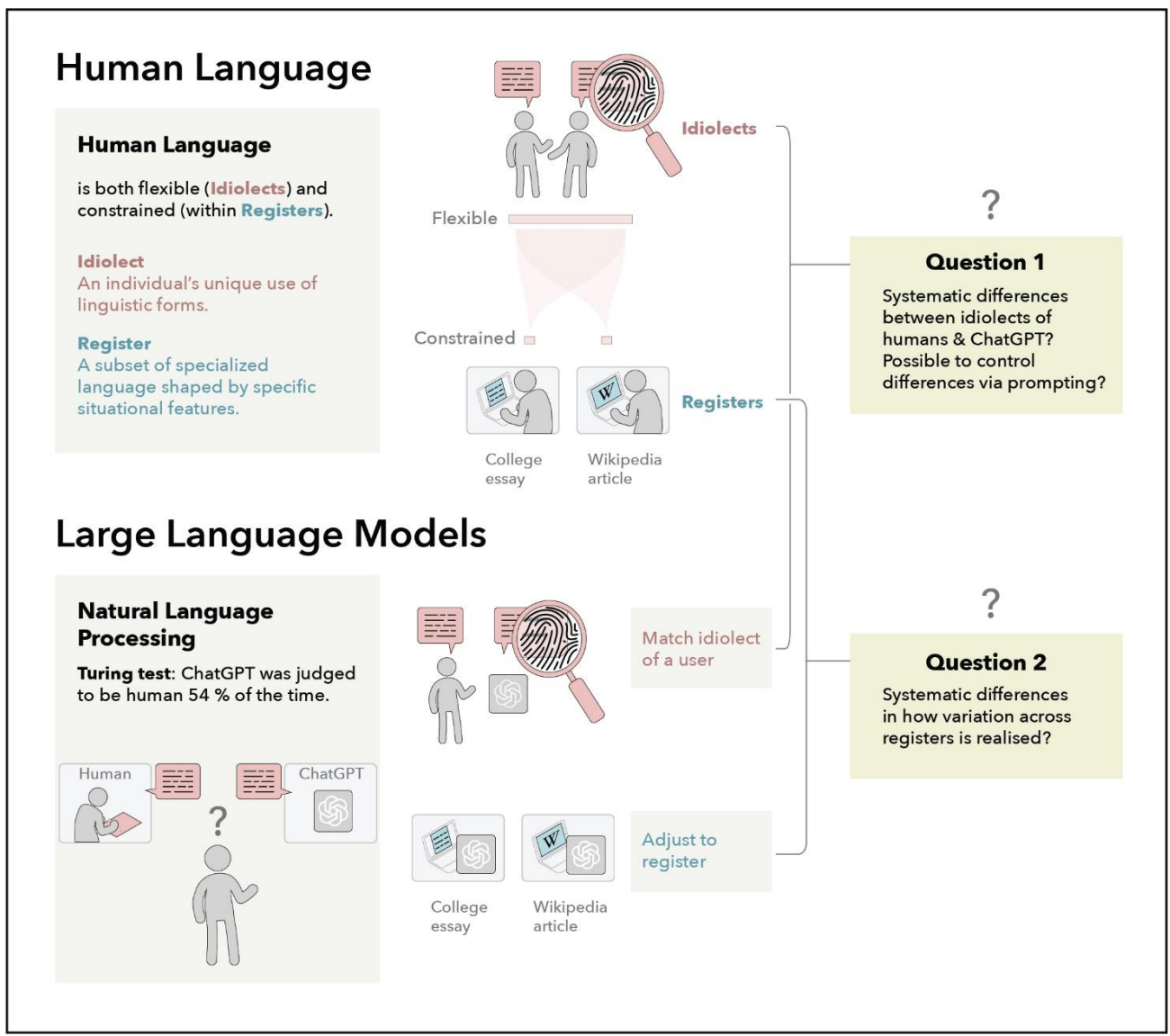

**Figure 1**. Study design and main questions.

**Results**

*Main Prompt Analysis*

A standard stylometric analysis is run to investigate the presence of any systematic stylistic differences between 2 sets of human texts and 4 sets of ChatGPT texts. Human texts are either (i) college argumentative essays (n=50) or (ii) Wikipedia encyclopedic entries (n=50). ChatGPT was prompted to generate college argumentative essays matched to (i) for topic (n=100; n=50 via a first prompt P1, n=50 via a second prompt P2), and to generate Wikipedia entries matched to (ii) for topic (n=100; n=50 via P1, n=50 via P2). We implement the analysis using the Stylo library in R [Eder et al. 2016].

Specifically, a FW-PCA is run [Grieve 2023], which identifies patterns of variation in function word frequencies. The FW-PCA is carried out on a feature set obtained by interrogating the entire corpus, from which we first extract the 100 most frequent words, that is, case-insensitive strings of alphabetic characters. As is standard in stylometry, we then concentrate on members of closed word classes, not only because they are frequent, thereby allowing for meaningful measurements to be made of their typical rates of use, but because they primarily express grammatical rather than topical information and are thus more likely to reflect patterns of authorial style across texts on different topics. We thus extract the first 27 most frequent words, which all occur at least twice per 1,000 words on average. Notably, after this point we also begin to get substantial numbers of content words (e.g., *life* at the 30$^{th}$ position). Accordingly, our selection does not feature nouns, adjectives, lexical adverbs, and lexical verbs, although auxiliary verbs and negation are included. The most common function word in the corpus is *the*, with a normalized frequency (i.e. frequency per every 1,000 words) of 29 [Figure 2, Upper panel]. Table 1 lists the 27 function words.

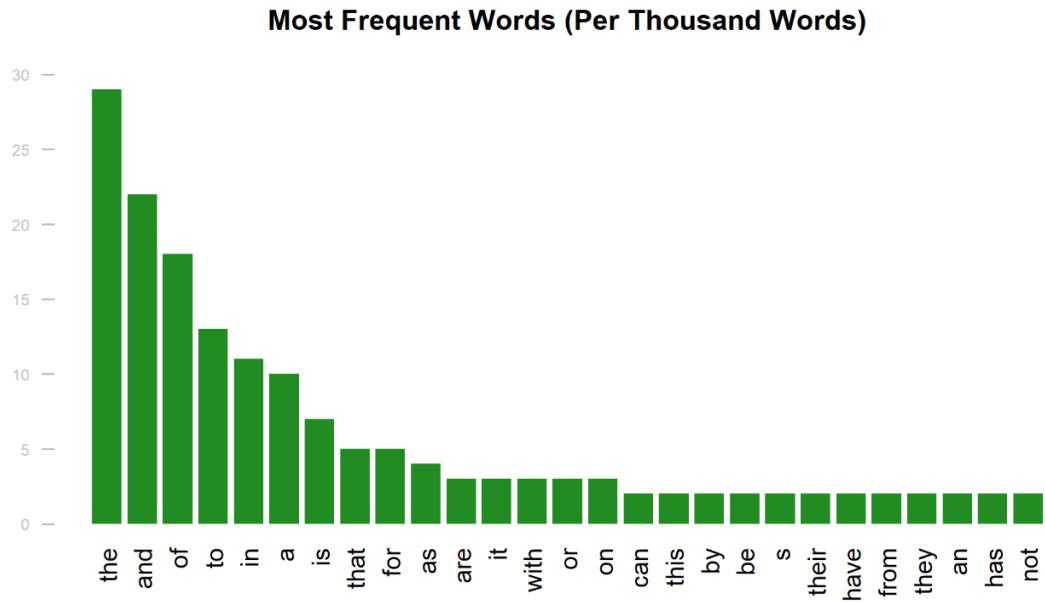

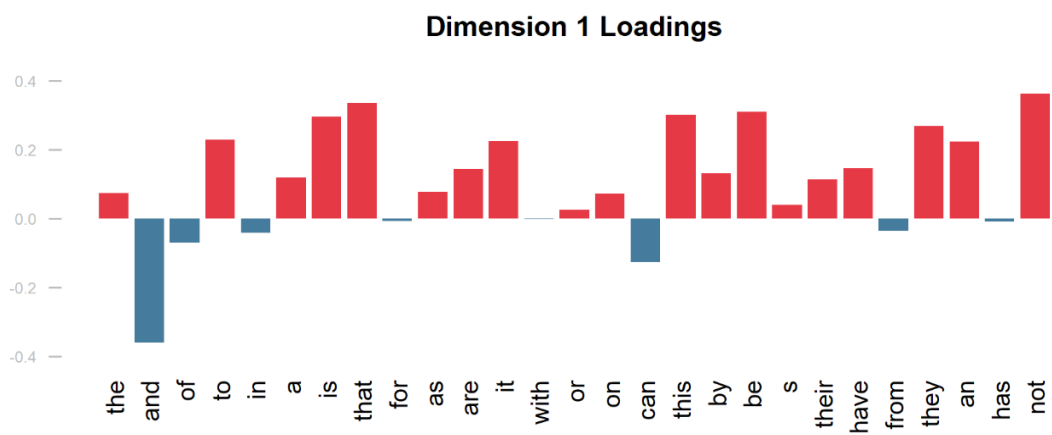

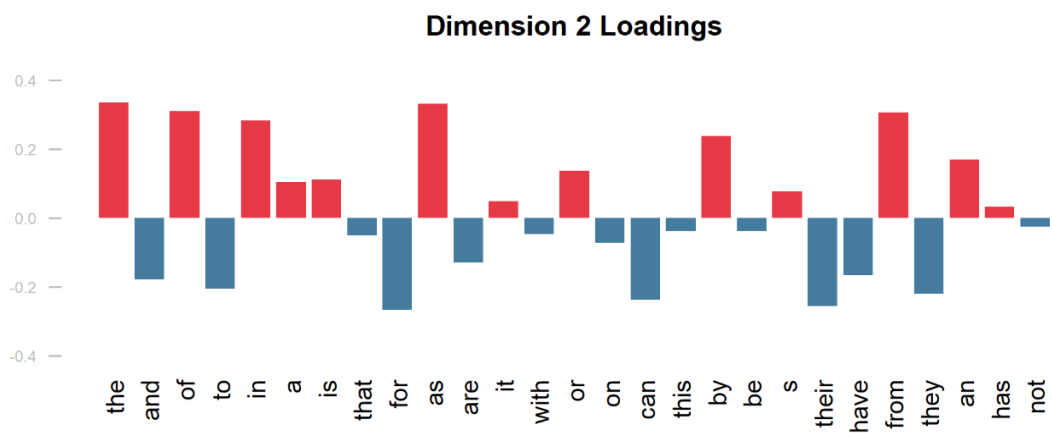

**Figure 2**. (Upper panel): Normalized frequency (i.e. frequency per every 1,000 words) for the FW-PCA feature set. (Middle panel): D1 loadings. (Lower panel): D2 loadings.

| | | | |
|---|---|---|---|
| the | and | of | to |
| in | a | is | that |
| for | as | are | it |
| with | or | on | can |
| this | by | be | s |
| their | have | from | they |
| an | has | not | |

**Table 1**: FW-PCA feature set.

To maintain a balance between the number of human- and model-authored texts under comparison, the feature set is subjected to 3 separate FW-PCAs. In **Stylometric Analysis on Prompt 2** and in **Stylometric Analysis on Prompt 1**, the human corpus is compared to ChatGPT texts generated via P2 and P1, respectively. Conveniently, the FW-PCA feature sets retrieved from the two sub-corpora are identical to the set in Table 1 for both analyses. Third, setting aside imbalances in the number of human- vs. model-authored texts and aiming to obtain the full picture, in **Additional Prompt Analysis** all human texts are compared to all ChatGPT texts. Last, a **Part-of-Speech Analysis** identifies differences in the usage of word classes.

**Stylometric Analysis on Prompt 2.** FW-PCA scores for 200 texts (i.e. 100 human and 100 ChatGPT texts) are presented [Figure 3, Upper panel]. Dimension 1 (D1) shows a clear difference between human texts (generally assigned positive scores) and ChatGPT texts (generally assigned negative scores), with human essays being especially distinctive ($W = 9233$, $p < .001$) [Figure 3, Middle panel]. A strong difference is also found within human texts (essays have positive scores, Wikipedia entries have negative scores; $W = 2438$, $p < .001$), whereas ChatGPT texts appear indistinguishable ($W = 1307$, $p = .697$). Given these prompts, ChatGPT failed to reproduce the clear pattern of register variation that can be seen when comparing human essays and encyclopedic writing. Overall, we find that ChatGPT writings of both types are more similar to human Wikipedia entries than to human essays.

Dimension 2 (D2) also shows a clear difference between human essays (negative scores) and human Wikipedia entries (positive scores) ($W = 200$, $p < .001$) [Figure 3, Lower panel]. Stylometrically, these two types of human writings are thus highly distinctive on both D1 and D2, as expected of different writing varieties [Biber & Conrad 2019]. Unlike D1, D2 shows a weak parallel trend within ChatGPT's two sets of texts ($W = 1000$, $p = .085$), indicating that ChatGPT minimally reproduces register differences. Lastly, for D2, humans generally obtain more positive scores, especially when compared to ChatGPT on Wikipedia entries ($W = 2284$, $p < .001$). Overall, D1 primarily splits human and ChatGPT essays, whereas D2 primarily splits human and ChatGPT Wikipedia entries.

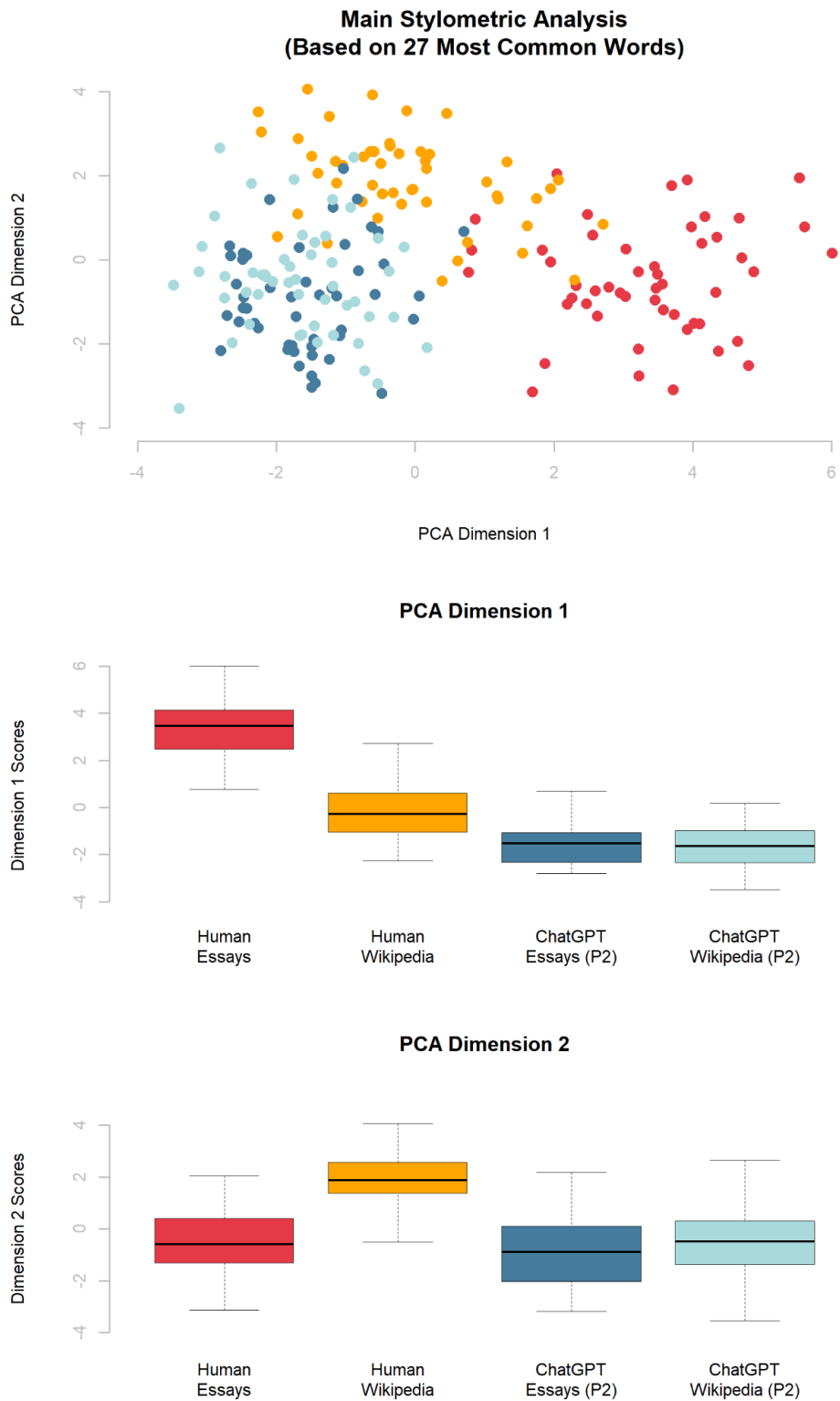

**Figure 3**. FW-PCA scores. (Upper panel): D1 and D2. (Middle panel): D1. (Lower panel): D2.

Next, we compare Interquartile Ranges across dimensions, that is, we measure the distance between the top and the bottom of the boxplots [Figure 3, Middle and Lower panels] ignoring whiskers. These results are presented in (1), which show that stylistic variation for human written texts is comparable to the amount of stylistic variation observed for model-generated texts. This is an unexpected outcome, considering that multiple humans authored the corpus texts, as opposed to one only LLM.

(1)
Human Essays: **D1** 1.621; **D2** 1.633
Human Wikipedia: **D1** 1.545; **D2** 1.185
ChatGPT Essays P2: **D1** 1.233; **D2** 2.074
ChatGPT Wikipedia P2: **D1** 1.286; **D2** 1.617

Lastly, factor loadings make explicit which words characterize each one of the poles of D1 and D2 [Figure 2, Middle and Lower panels]. Table 2 summarizes the strongest D1 and D2 loadings. These results indicate that ChatGPT's texts are generally characterized by the presence of fewer function words, with the exception of the conjunction *and*, which stands out as a characteristic of all ChatGPT texts. This preliminary observation will be interpreted in more detail through the MDA that follows.

|  | Dimension 1 | Dimension 2 |
|---|---|---|
| **Positive loadings** | not (.36) | as (.33) |
|  | that (.33) | the (.33) |
|  | be (.31) | from (.31) |
|  | this (.30) | of (.31) |
|  | is (.30) | in (.28) |
|  | they (.27) |  |
| **Negative loadings** | and (-.36) | for (-.27) |
|  |  | their (-.26) |

**Table 2**: Factor loadings for D1 and D2. D1 positive loadings are associated with human texts, especially essays. The only D1 negative loading is associated with all ChatGPT texts. D2 positive loadings are associated with Wikipedia entries, especially human-authored. D2 negative loadings are associated with essays, especially ChatGPT-generated.

To summarize, in terms of the main aggregated patterns of stylometric variation, (i) human and ChatGPT writings are very different, and (ii) human writing exhibits far more register variation than ChatGPT's writing, where very limited register variation is noted.

**Stylometric Analysis on Prompt 1.** To explore how much the results presented in the first FW-PCA are sensitive to variation in prompting, we subject an additional set of ChatGPT texts to another FW-PCA. Scores for 200 texts (i.e. 100 human and 100 ChatGPT texts) are presented [Figure 4, Upper panels].

The amount of variance and the overall results mirror the patterns found in the first FW-PCA. However, notably, more register variation is found in ChatGPT texts generated using this set of prompts. This is especially evident on D1, where we find that ChatGPT essays are very close in similarity to human Wikipedia entries (whereas ChatGPT Wikipedia texts are found at the more extreme end of D1) [Figure 4, Upper left panel].

For this set of prompts, we find that the strength of the difference within ChatGPT text varieties is now non-negligible on D1 ($W = 2328$, $p < .001$). This indicates that the employment of different prompts influences variation, in this specific case by increasing it. However, because the degree of register variation exhibited by ChatGPT still does not match the degree of register variation found across human texts (D1: $W = 2443$, $p < .001$; D2: $W = 157$, $p < .001$), it is clear that ChatGPT is still struggling to replicate natural register variation. On D2, significant differences within ChatGPT varieties are present ($W = 1014$, $p = .032$) [Figure 4, Upper right panel].

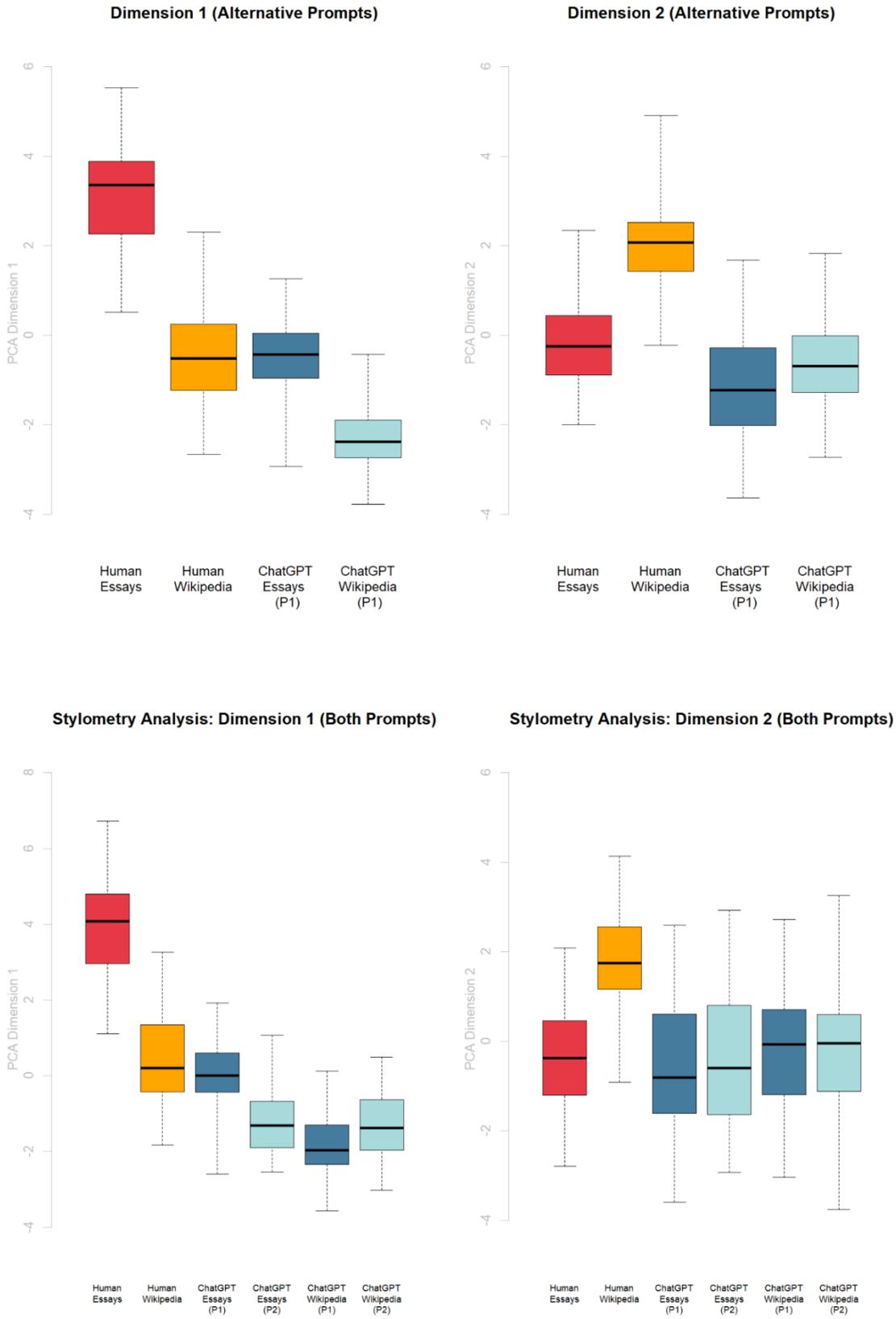

**Figure 4**. FW-PCA scores on texts generated via P1 (Upper panels), and on texts generated via both prompts (Lower panels).

**Additional Prompt Analysis.** To better understand the relationship between these different prompting strategies, we subject the full corpus —and thus the same feature set in Table 1— to a third FW-PCA [Figure 4, Lower panels]. We clearly see that all AI-written texts are much more similar to each other on both dimensions than they are to human texts, which show far more pronounced patterns of register variation, regardless of which prompt is used. Overall, these results offer further support for the two main conclusions drawn from the first two FW-PCAs, namely: (i) human texts are clearly distinguishable from ChatGPT texts, and (ii) human texts show substantially more register variation than ChatGPT texts.

**Part-of-Speech Analysis.** In this analysis, we build on the previous results and investigate hypotheses concerning the presence (or absence) of particular linguistic features in the corpus. We do so by focusing on ChatGPT texts generated via P2 (n=100). The dimension loadings in the stylometric analysis on P2 show that ChatGPT texts are overall low on function words —with the notable exception of *and*—, and may thus be high on content words, especially nouns and adjectives, as the MDA analysis below confirms. The hypothesis is that ChatGPT texts could primarily feature dense noun phrases, characterized by long sequences of nouns being modified, especially pre-nominally, both by adjectives and other nouns, at times coordinated with the conjunction *and*. The presence of this pattern, which is common in informational texts [Biber 1988], can be verified through an analysis of word classes.

Nouns, and especially nominalizations, are highly characteristic of ChatGPT-authored texts [Figure 5]. By looking at adjectives, we find a high number of attributive constructions and a comparatively lower number of predicative constructions: nouns are internally modified (i.e. pre-nominally), as opposed to being modified through copular constructions. Additionally, by looking at *and*, we find predominance of phrasal coordination in ChatGPT as opposed to human data. Relatedly, there are certain word classes expected to be low in the ChatGPT texts in light of previous results: verbs, adverbs and pronouns. The distinctively low frequencies for past tense verbs and adverbs in ChatGPT data are especially notable.

In sum, ChatGPT texts showcase a high count of nouns, nominalizations, attributive adjectives, and phrasal coordination, and low counts for past tense verbs, adverbs, third person pronouns, and predicative adjectives. These results are in line with our previous analyses and provide the basis for the development of a more interpretable explanation for these results.

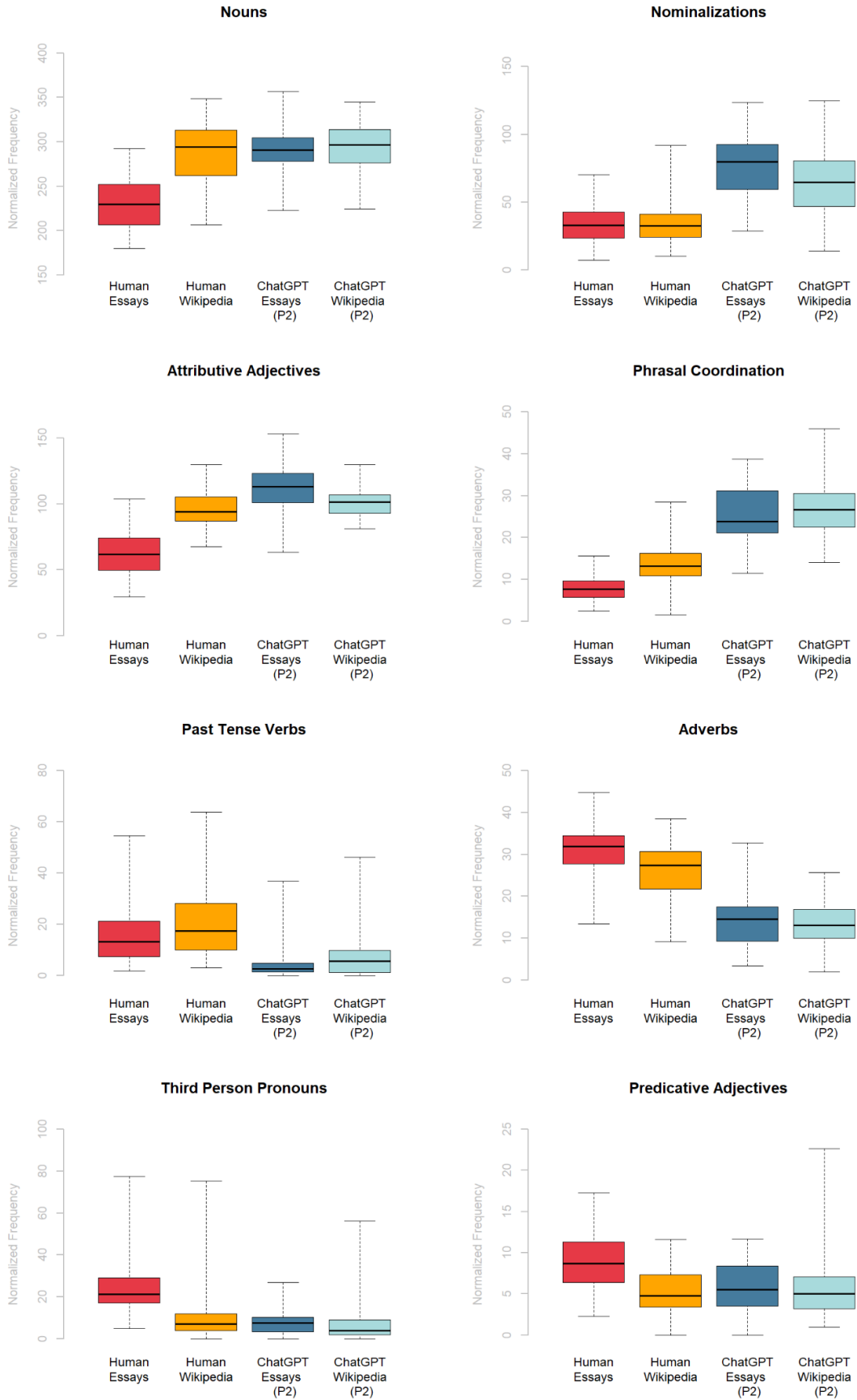

**Figure 5**. Normalized frequency (i.e. frequency per every 1,000 words) for 8 individual word-class tags.

*Multidimensional Register Analysis*

Lastly, an MDA is conducted following the method outlined in [Biber 1988, see Methods] using the Multidimensional Analysis Tagger [Nini 2019]. This type of analysis allows us to interpret the results obtained with stylometry [Grieve 2023]. With an MDA, new dimensions of register variation are calculated based on a selected feature set, focusing on part-of-speech tags that are most common across the investigated texts (n=200, with P2 texts). To focus our analysis on frequent features, we select tags that occur at least 5 times per 1,000 words. This results in the selection of 20 variables (Table 3).

| ANDC Clausal Coordination | X.BEMA Be as Main Verb | DEMO Demonstratives | GER Gerunds |
|---|---|---|---|
| JJ Attributive Adjectives | NN Nouns | NOMZ Nominalizations | X.PASS Passives |
| X.PEAS Perfect Aspect | PHC Phrasal Coordination | PIN Prepositional Phrases | PIT Pronoun It |
| POMD Possibility Modals | PRED Predicative Adjectives | X.PRIV Privative Verbs | RB Adverbs |
| TO To Infinitives | TPP3 Third Person Pronouns | VBD Past Tense | VPRT Present Tense |

**Table 3**: MDA feature set.

The MDA feature set is first reduced to 3 dimensions using factor analysis. The choice to reduce to 3 dimensions follows from the observation that these account to approximately 50% of the variance, while adding a 4th dimension does not account for substantially more variance. Crucially, these first 3 dimensions are also readily interpretable from a stylistic perspective. The results are shown in [Figure 6], while the loadings associated to each dimension are presented in Table 4.

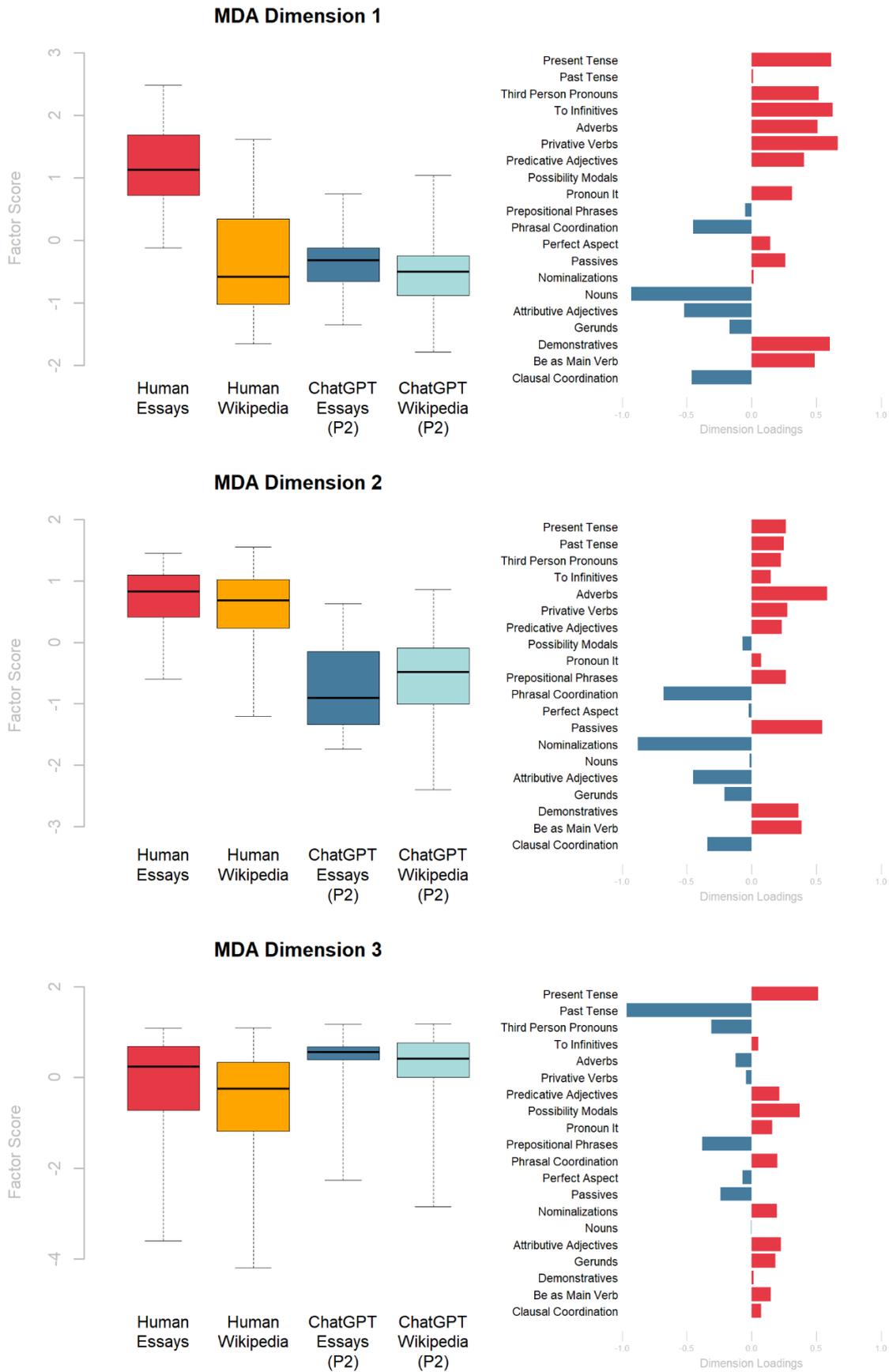

**Figure 6**. MDA factor scores and dimension loadings for D1 (Upper panels), D2 (Middle panels), and D3 (Lower panels).

D1 represents an information density pattern. This is in line with previous results in MDA research, where an opposition between more and less informationally dense style is often reflected in the first dimension [Biber 1988]. Specifically, we find that human essays are less informationally dense, with less complex noun phrases and with substantially higher loadings for pronominals and verbal auxiliary forms. Human essays are also clearly differentiated from human Wikipedia entries, whose style is largely mirrored in both types of ChatGPT texts. In other words, on D1 ChatGPT levels out register differences in its outputs [Figure 6, Upper panels].

Alternatively, D2 identifies a clear difference between human vs. ChatGPT texts, identifying relatively limited internal register variation within either of these two corpora [Figure 6, Middle panels]. ChatGPT texts are characterized by a maximization of nouns and adjectives over all other word classes. This clear difference from human writing extends our basic stylometric findings, confirming that ChatGPT features in its outputs mainly nominalizations and noun phrase-internal coordination. ChatGPT also avoids the use of adverbs and passives, both of which are generally optional, omittable or replaceable with active sentences, and often prescribed against in formal writing, with consequences on function word frequencies. Unlike the other dimensions, this is a relatively unique pattern, not regularly attested in previous MDA research, presumably reflecting the unique characteristics of model-generated texts.

Lastly, D3 identifies a pattern of stylistic variation related to the production of narrative texts, which is once again well attested in previous MDA research [Biber 1988], where narrative texts are characterized by the relatively frequent use of past tense and third person pronouns, and where non-narrative texts show a complementary pattern, including the frequent use of present tense. In this case, the analysis finds that ChatGPT disfavors narration and, once again, exhibits less register variation than humans, as reflected by substantially less variance in the text-level factor scores for AI-generated texts on this dimension [Figure 6, Lower panels].

|  | Dimension 1 | Dimension 2 | Dimension 3 |
|---|---|---|---|
| **Positive loadings** | Privative Verbs (.66) <br> To Infinitives (.63) <br> Present Tense (.62) <br> Demonstratives (.60) <br> Third Person Pronouns (.52) <br> Adverbs (.51) <br> Be as Main Verb (.49) <br> Predicative Adjectives (.41) | Adverbs (.59) <br> Passives (.54) | Present tense (.51) |
| **Negative loadings** | Nouns (-.93) <br> Attributive Adjectives (-.52) | Nominalizations (-.88) <br> Phrasal Coordination (-.68) | Past tense (-.97) <br> Third Person Pronouns (-.31) |

Clausal Coordination (-.46)
Phrasal Coordination (-.45)

**Table 4**: MDA factor loadings for D1, D2 and D3. D1 positive loadings are associated with human essays. D1 negative loadings are associated with human Wikipedia entries and with ChatGPT texts. D2 positive loadings are associated with all human-authored texts. D2 negative loadings are associated with all ChatGPT texts. D3 positive loadings are associated to all texts in the corpus. D3 negative loadings are associated with human texts.

Finally, as we have seen throughout both sets of analyses, in general, it seems as if AI-generated texts are characterized by more frequent use of content words, whereas human generated texts are characterized by more frequent use of function words. Drawing on the tagged corpus used in the second set of analyses presented here, we test this hypothesis directly, by comparing the frequencies of open and closed word classes together. To this end, a function vs. content word analysis compares the proportion of function and content word frequencies in human vs. ChatGPT texts. This analysis is added in light of the striking differences found in the preliminary scrutiny that allowed us to extract the data to subject to the FW-PCAs. In the top 50 most frequent words in the human and ChatGPT P2 corpus (n=200 texts), in fact, figure 8 content words: *life*, *ethical, public, human, education*, *water*, *people* and *students*. Typically, fewer content words figure in this count (e.g., 2 content words in the top 50 most frequent words for the human-written corpus analyzed in [Grieve 2023], which are usually removed from subsequent analyses.

To gain insight as per the origin of such high content-word count, the human corpus (n=100 texts) and the ChatGPT P2 corpus (n=100) are analyzed separately, and two lists of the top 50 most frequent words in each are built. In the human data, *people* is the first content word, and it figures at the 32$^{nd}$ position. The only other content word in the top 50 is *other*, which ranks 40$^{th}$. The next, *water*, comes in at 61$^{st}$. Notably, both of these words are also relatively generic, in the sense that they do not reveal much about the content of a text. Alternatively, in ChatGPT's data, in the top 50 words we get 18 content words (i.e. *ethical*, *education*, *life*, *human*, *individuals*, *legal*, *public*, *challenges*, *water*, *health*, *cultural*, *impact*, *social*, *role*, *historical*, *various*, *society*, *schools*). The first content word *ethical* comes in at the 19$^{th}$ position, which, along with many of these words, is relatively specific.

This signals a remarkable difference between humans and ChatGPT: ChatGPT makes use of a much higher proportion of content words, especially nouns and adjectives. This finding points to a very informationally dense writing style —albeit an unusual one, for it does not match human informationally dense texts [Biber 1988].

**Discussion**

This work is guided by 2 RQs: RQ1. Are there systematic stylistic differences between humans and ChatGPT? How much can these differences be controlled via prompting? RQ2. Are there systematic stylistic differences between humans and ChatGPT in how variation across registers is realized?

In relation to RQ1, we find clear and describable differences between humans and ChatGPT across all our analyses. Some of the observed differences can be marginally mitigated via prompting (and presumably further via model pre-training and fine-tuning targeting register variation, as is common in domain adaptation [Gururangan et al. 2020, Grieve et al. 2025]) —a point that raises *stimulus-hacking* concerns when addressing the question of whether AI has passed the Turing Test, to which we return below. In terms of describing ChatGPT's writing style, it is highly informationally dense, marked by an extensive use of all-things nominal: nouns, nominalizations, and attributive adjectives. Consequently, ChatGPT appears to be linked to a unique writing style that is characterized by the prominent role of the nominal domain. This comes at the expense of a limited use of pronouns, adverbs, and (passive) verbs, as well as function words in general, with the exception of *and*, which is used to coordinate structures, especially at the phrasal, as opposed to the clausal, level. Alternatively, we find that human essays are less informationally dense.

This difference in the writing style of humans and machines appears to occur because ChatGPT is an artificial agent that lacks both communicative needs of its own and the ability to read the communicative intents and needs of its interlocutors. Thus, for humans, information density is proportional to our expectations about how our linguistic message will be understood by our interlocutors —which is something that LLMs cannot assess, as they are generally not trained to generate language across ranges of registers. The tested model does not experience the full range of human registers, especially not in the balance that we do, nor is it really expected to engage in the full range of these registers, especially not in more interpersonal, interactive, social uses, where the highly efficient expression of detailed information is not the primary driver of communication.

Furthermore, it is possible that ChatGPT's writing style/idiolect is different from that of humans because it is less anchored in the highly grammaticalized dimensions of tense, aspect, mood, and the other functional categories of what linguists call the fine and highly articulated structure of the *left periphery* [Rizzi 1997]. To explain this, syntactic representations are complex objects consisting of sequences of hierarchically organized functional chunks of language. The left periphery refers to the topmost portion of the syntactic representation that is situated above the Inflectional Phrase (i.e. the syntactic position for encoding verb inflection), and that expresses, among other things, tense, aspect, mood, topic, focus, and illocutionary force. It is perhaps unsurprising that the tested model avoids precisely the top parts of the verbal domain because this is the portion of grammar that grounds an utterance into the discourse and the world. A noun like *snow* or a proper name like *John Kennedy* cannot be assessed in terms of their status as true or false. For that we need to add the ingredients of the verbal domain: 'snow *was falling* in Nebraska on January 1, 2023' and 'John Kennedy *is* dead' are both true. LLMs struggle with truth conditions and do not provide a good basis for truth [Marcus 2001, 2024]. Their responses may be plausible, confident, and largely correct, but may still contain factual inaccuracies [Wachter et al. 2024]. Precisely because models cannot assess the real-world conditions that render a statement true or false, it is unsurprising that they avoid the pronounced use of linguistic devices that expose this weakness. In an interesting parallel with human development, the left periphery is also the one that comes in last and leaves first: It is acquired last during first language development [Friedmann et al. 2021] and is fragile and less likely to be spared in cases of brain damage (e.g., agrammatic aphasia [Friedmann 2002]). Overall, the more complex aspects of grammar, which is the backbone of language and has been argued to reflect the organization of a specific mode of thought that is unique to humans [Hinzen 2013], are the ones that allow for a clearer manifestation of the differences in the writing style of humans vs. the tested model.

Our second finding, addressing RQ2 and whether there are differences between humans and ChatGPT in how variation across registers is realized, is that ChatGPT shows limited variation when prompted to produce texts in different registers. While the tested model does not completely ignore register variation, it fails to replicate it in a human-like manner, nor does it produce register variation by default, as is true of human language; rather, it requires careful and explicit prompting to generate stylistic patterns that are somewhat sensitive to variation in communicative context. This inability probably stems both from the previously noted limitations in grasping the nuanced uses of the most complex domains of grammar (in the sense that there is less room for adapting) and from a lack of knowledge of the pragmatic requirements associated with the notion of register. Put another way, the AI style we have identified through the tested model seems to be less flexible, less adaptive, and less context-sensitive than its human-authored counterparts. It seems that although AI can adapt for register, there are some engineered characteristics that may help us distinguish its outputs from human outputs, thus aiding in authorship identification. This finding can be interpreted through the notion of *domain adaptation*: The degree of the models' ability to adapt for register also depends on the relationship between the larger variety represented by the training material and the smaller target variety towards which the model is being prompted or adapted [Grieve et al. 2024].

In sum, although we find that the tested model can adapt for register in a limited manner, this result is at odds with recent claims that LLMs can adapt so well that their conversational output is passed for human more often than not [Jones & Bergen 2024]. We argue that this difference in results reflects methodological choices. Concretely, our prompts neither identify a very small and clearly delineated target variety, nor are they the outcome of a pre-testing session of various prompts, and a subsequent selection of those that give the clearest results. In other work, different prompts were pre-tested, and the main experiment was performed only after fine-tuning and enriching the best-performing ones: "In a prior, exploratory study [...] we tested a wide variety of different prompts—varying the personality, strategy, and linguistic style they instructed the model to adopt. In the present work, we adapted the best performing prompt from the exploratory study [...]" and "The first part of the prompt instructs the model to behave as if it is a young person who is not taking the game too seriously, uses some slang, and makes spelling and grammar errors" [Jones & Bergen 2024]. Put another way, if we ask the model to write rap lyrics in English and provide it with some slang to use in its responses, this already identifies a very concise and clearly demarcated variety in a way that aids domain adaptation. This is very different from asking the model to adapt for register by producing student college essays vs. Wikipedia entries. In our experimental setting, the boundaries of the target varieties are not so clear, and these subtle differences may make adapting for register harder. Human-authored texts can capture such differences, because humans have linguistic knowledge that is developed and rooted in real-world conditions. Pre-testing and fine-tuning the prompts is undeniably useful for exploratory analyses, but it also raises concerns for s(timulus)-hacking. In p-hacking, different analyses are explored until a result is achieved. In s-hacking, different stimuli/prompts are explored until a result is achieved. Overall, evaluating the ability of LLMs to adapt for register is sensitive both to methodological tweaking and to the sociolinguistic dimensions of the target variety. Observing that LLMs can variably adapt for register depending on the methodology brings again to the fore Lady Lovelace's Objection: AI can do whatever we know how to order it to perform.

**Methods**

In this section we introduce the corpus data and present our choice of analyses.

*Data*

The corpus analyzed in this work consists of 300 texts. These texts are divided into 6 groups: 2 authored by humans and 4 authored by ChatGPT. The 2 human groups are texts written in two registers: (i) **Human Essays**. These are human single-authored texts (n=50, totaling 61,029 words, with a mean length of 1,221 words). Each text has one college student author who is a native speaker of English. The texts are retrieved from the Locness Corpus [Granger 1998]. (ii) **Human Wikipedia**. These are human multi-authored texts (n=50, totaling 59,326 words, with a mean length of 1,187 words). Each text corresponds to the initial paragraphs of a Wikipedia entry on a given topic, chosen as a general source of variation. Wikipedia pages were accessed in November 2023.

The 4 ChatGPT groups of texts are obtained via prompting ChatGPT to generate texts matching Human Essays and Human Wikipedia on the same topics. For exploratory purposes, two different prompting strategies are used without any pre-testing: this permits to assess the effect of different prompts on our results.

The first set of prompts (P1) used for ChatGPT does not specify a desired output length. Having however noticed that ChatGPT's outputs tended to be at the lower limit of what is considered an unproblematic length in stylometry (outputs averaged 672 words), in a second set of prompts (P2) ChatGPT was asked for 1,000-word texts. While it is not the case that ChatGPT provided us with 1,000-word outputs, these were nonetheless considerably longer (with the mean length of ChatGPT texts in number of words being 924) than default outputs. The content of the prompt request remained unvaried, only wording changed. The texts generated via P2 were collected chronologically after those generated through P1. However, since longer texts are more suitable for stylometric investigations, we first present the results on longer texts (obtained through P2), and then the results obtained through P1.

The analysis on the prompts we present is intended to introduce a methodological baseline by investigating *whether* register differences can be modulated via different prompts, as opposed to *how* register variation is affected by different prompting strategies. For that, more systematic testing is needed. Four additional groups of texts thus constitute our corpus: (iii) **ChatGPT Essays P1**. These are model-generated texts (n=50, totaling 30,990 words, with a mean length of 920 words) on the same topic as Human Essays, elicited using the prompt: "Write a college argumentative essay on *topic*". (iv) **ChatGPT Wikipedia P1**. These are model-generated texts (n=50, totaling 36,230 words, with a mean length of 725 words) on the same topic as Human Wikipedia, elicited using the prompt: "Write a Wikipedia article on *topic*". (v) **ChatGPT Essays P2**. These are model-generated texts (n=50, totaling 41,178 words, with a mean length of 824 words) on the same topic as Human Essays, elicited using the prompt: "Pretend to be a college student and write a 1,000-word essay on *topic*". (vi) **ChatGPT Wikipedia P2**. These are model-generated texts (n=50, totaling 51,128 words, with a mean length of 1,023 words) on the same topic as Human Wikipedia, elicited using the prompt: "Write a 1,000-word encyclopedia page on *topic*". Summing up, there are 100 human-authored texts (120,355 total words, mean length of 1,204 words) and 200 ChatGPT texts (159,526 total words, mean length of 798 words). The full list of topics, together with the corpus of the analyzed texts and the annotated code employed for the analyses and for plotting can be found at https://osf.io/w5pgr/?view_only=f2dc4b9493374f3b88413c63083ace70. Queries were put into the ChatGPT-3.5 application programming interface, set on default parameters, in November 2023. ChatGPT-3.5 is a fine-tuned version of GPT-3.5 [Ouyang et al. 2022], further integrated by reinforcement learning from human feedback. This model was chosen for being representative of the state-of-the-art at the time of data collection, and for its centrality in the LLM ecosystem [Bommasani et al. 2024].

*Analyses*

The joint employment of stylometry and multidimensional techniques has been established as a methodological baseline in [Grieve 2023]. It has the purpose of (i) providing a holistic view of stylistic variation between groups of texts, by allowing the identification of basic variation patterns; and (ii) providing a firm line for the interpretation of the observed patterns. In addition to analyzing function word frequencies, we also examine Part-of-Speech (POS) frequencies, which is both common in stylometry [Baayen et al. 1996] and the basis for MDA [Biber 1988]. Notably, although the multivariate analysis of function word frequencies and POS tags tend to produce comparable results [Grieve 2023], the analysis of POS, especially from an MDA perspective, tends to provide a better basis from which to describe and explain these underlying patterns of stylistic variation in a holistic and insightful manner.

**Stylometry.** Stylometry is operationalized in this work as a Function Word Principal Component Analysis (FW-PCA) and is a common and well-established method in stylometry [Binongo 2003, Eder 2016, Grieve 2023]. The FW-PCA approach is a multivariate method for dimension reduction (for a related method that generally produces comparable results see Burrow's Delta [Burrows 2002]): given the relative frequency of common function words measured across texts in a corpus, the FW-PCA projects the texts into low-dimensional space and allows scrutiny of the aggregated dimensions. First, a correlation matrix is calculated so that each pair of linguistic variables in the data across all texts is assigned a correlation measure. Second, aggregated dimensions are extracted from the correlation matrix, where each of these dimensions accounts for a decreasing amount of variance.

Each dimension is then associated with *dimension loadings*. The magnitude of the loadings reflects how strongly each linguistic variable is represented by a given dimension. The sign (positive or negative) of the loadings specifies what variables are either positively or negatively correlated with each other: variables that are positively correlated (i.e. that tend to occur together in texts) are assigned the same sign, whereas variables that are negatively correlated (i.e. that do not tend to occur together in texts) are assigned opposing signs.

Every dimension is further associated with *dimension scores*, which specifies the degree to which each text is characterized to that pattern of linguistic variation: texts assigned strong positive scores contain frequent use of the variables assigned positive loadings, and infrequent use of the variables assigned negative loadings, whereas texts assigned strong negative scores contain frequent use of the variables assigned negative loadings and infrequent use of the variables assigned positive loadings.

By proceeding in this way, we move from a data matrix where every corpus text is defined by a large number of individual linguistic variables, to a data matrix where every corpus text is defined by a reduced number of aggregated variables, each accounting for the largest possible share of variation in the original data matrix and each representing a distinct pattern of stylistic variation. Up to this stage of analysis, the source of the texts is not disambiguated. Once the dimensions are extracted, however, the different texts are compared based on dimension scores with the objective of identifying the dimensions that most accurately distinguish amongst styles of writing. Usually, only one dimension strongly disambiguates between the sources, while the other dimensions represent other sources of variation.

**Multidimensional Register Analysis.** MDA is a quantitative method for linguistic analysis developed by [Biber 1988], which is related to stylometry, as it is based on similar methods. However, as opposed to stylometry (which is generally used to disambiguate the authors of different groups of texts), MDA is generally used to compare and explain linguistic variation across registers, for instance, conversations, novels, lectures, etc. As is the case in stylometry, here too the relative frequencies of grammatical forms are measured across texts, which are subsequently subjected to multivariate statistical analysis with the purpose of identifying the main aggregated dimensions of linguistic variation that differentiate such texts. Crucially, these dimensions are interpreted as dimensions of *register* variation: the linguistic features associated to each dimension are interpreted as representing differences in the communicative purposes of different types of texts (e.g., the frequent use of personal pronouns in personal conversations is interpreted as being related to the general need to repeatedly reference people in this communicative context).

MDA provides a theoretical foundation for explaining stylometry results: if differences between two authors are found, these are hypothesized to be rooted in the fact that these authors express themselves in subtly different registers [Grieve 2023]. Particularly, the differences in register identified through an MDA signal awareness to context, as situational components play a role in determining how we write. Since our corpus texts involve very different situational components, they are expected to showcase features which adapt to the register of each context (e.g., an essay as opposed to an encyclopedia entry).

The MDA conducted in this work follows the method outlined in [Biber 1988], whereby texts written in similar styles get clustered together. Biber [1988] first identified a set of 67 analyzable grammatical features that he then reduced to 6 dimensions of linguistic variation. For instance, the first of these dimensions contrasts texts on their information density: formal writings tend to feature a high count of nouns and nominal modifiers, whereas more involved and spontaneous genres are more likely to feature pronouns and a large number of verbs. The second dimension instead typically contrasts narrative vs. non narrative texts (the former containing lots of verbs in the past tense).

By selecting 20 features of the 67 that were first proposed in Biber [1988] (we made our selection as a function of the high frequency of such features), we identified 3 main dimensions of variation and interpreted them following the registers identified in Biber [1988]. The conversion of the base corpus to its grammatically-tagged version (from which MDA features were extracted) was operated using the Multidimensional Analysis Tagger [Nini 2019], a software package based on the Stanford Tagger, which also generates relative frequency measures (per hundred words) for each feature for each text in the corpus, as required to perform the MDA.

Last, the results obtained in the present work are clear-cut, but limitations in our methodology should be noted. Our testing is limited to one LLM and two prompting strategies. While we do not claim that our results fully generalize to all LLMs as a family, we maintain that we have established a novel methodological baseline for the characterization and detection of AI-generated content. In future work, this analysis can be expanded both in terms of registers and LLMs, in order to allow increasing insights into the content-generation strategies that are behind LLMs.

**Data and Code Availability**

The full human-authored and ChatGPT-generated corpora, together with the annotated code employed for the analyses and plotting are available at
https://osf.io/w5pgr/?view_only=f2dc4b9493374f3b88413c63083ace70.

**Acknowledgments**

We acknowledge the contribution of the Centre for English Corpus Linguistics, Université Catholique de Louvain, Belgium, in making the Locness corpus available for consultation.

**Author contributions**

V.D. and E.L. designed research and wrote the paper; V.D. collected data; W.H. and J.G. analyzed data; S.A.M. assisted with code editing. All authors contributed to the final version of this manuscript.

**Competing interests**

The authors declare no competing interests.

**Materials & Correspondence**

Correspondence and material requests should be addressed to Vittoria Dentella, Università degli Studi di Pavia, Department of Brain and Behavioral Sciences, Piazza Botta, 11, 27100 Pavia, Italy. Email: vittoria.dentella@unipv.it